\documentclass[conference]{IEEEtran}

\usepackage{subfigure}
\usepackage{graphicx}
\usepackage{algpseudocode}
\usepackage{algorithm}
\usepackage{amsmath}
\usepackage{multirow}

\ifCLASSINFOpdf
\else
\fi

\begin{document}
%
\title{Towards a Dedicated Computer Vision Toolset for Crowd Simulation Models}

\author{\IEEEauthorblockN{Sultan Daud Khan}
\IEEEauthorblockA{Makkah Techno Valley Company\\
Makkah, Saudi Arabia\\
 sdkhan@gistic.org}
\and
\IEEEauthorblockN{Muhammad Saqib,Michael Blumenstein}
\IEEEauthorblockA{Institute of Integrated and Intelligent Systems,\\
Griffith University\\
{muhammad.saqib,michael.blumenstein}@griffithuni.edu.au}
}

\maketitle

\begin{abstract}

As the population of world is increasing, and even more concentrated in urban areas, ensuring public safety is becoming a taunting job for security personnel and crowd managers. Mass events like sports, festivals, concerts, political gatherings attract thousand of people in a constraint environment, therefore adequate safety measures should be adopted. Despite safety measures, crowd disasters still occur frequently. Understanding underlying dynamics and behavior of crowd is becoming areas of interest for most of computer scientists. In recent years, researchers developed several models for understanding crowd dynamics. These models should be properly calibrated and validated by means of data acquired in the field. In this paper, we developed a computer vision tool set that can be helpful not only in initializing the crowd simulation models but can also validate the simulation results. The main features of proposed tool set are: (1) \emph{Crowd flow segmentation and crowd counting}, (2) \emph{Identifying source/sink location for understanding crowd behavior}, (3) \emph{Group detection and tracking in crowds}.

\end{abstract}


%
\IEEEpeerreviewmaketitle

\section{Introduction}

Crowds of pedestrians can be considered as complex entities from different points of view: the variety individual and collective behaviours that take place in a crowd, the composite mix of competition for the space shared by pedestrians but also the collaboration due to the not necessarily explicit but often shared (at least in a given scenario) social norms, the possibility to detect self-organization and emergent phenomena
they are all indicators of the intrinsic complexity of a crowd. The relevance of human behaviour, and especially of the movements of pedestrians, in built environment in normal and extraordinary situations (e.g. evacuation), and its implications for the activities of architects, designers and urban planners are apparent, especially considering dramatic episodes such as terrorist attacks, riots and fires, but also due to the growing issues in facing the organization and management of public events (ceremonies, races, carnivals, concerts,parties/social gatherings, and so on) and in designing naturally crowded places (e.g. stations, arenas, airports). The phenomena of crowd like sports, festivals, concerts, political gatherings etc, are mostly observed in urban areas, which attracts hundreds of thousands people. Pedestrian and crowd modelling research context regards events in which a large number of people may be gathered or bound to move in a limited area; this can lead to serious safety and security issues for the participants and the organisers. The understanding of the dynamics of large groups of people is very important in the design and management of any type of public events. In addition to safety and security concerns, also the comfort of event participants is another aim of the organisers and managers of crowd related events. Large people gatherings in public spaces (like pop-rock concerts or religious rites participation) represent scenarios in which crowd dynamics can be quite complex due to different factors (the large number and heterogeneity of participants, their interactions, their relationship with the performing artists and also exogenous factors like dangerous situations and any kind of different stimuli present in the environment. Such crowding phenomena poses serious challenges to public safety and crowd management. Therefore analysis of crowd is crucial for solving real world problems. Researchers from different communities like sociology, civil, physics and computer science are studying crowding phenomena from different angles. Besides these efforts, computer vision research community developing algorithms that can automatically understand the crowd dynamics in the real-world scenes. Despite these efforts, computer vision research community have not achieved the desired level of applicability and robustness. This is due to the following reasons; (1) The algorithms are based on particular assumptions which are often violated in real-world environment (2) Due to scarce data, computer vision models can not be trained effectively.

In order to improve computer vision algorithms, computer simulation have been used to provide the training data and also for validation of the algorithms.

\section{Crowd Dynamics: An Integrated Approach}

\begin{figure}[h]
\includegraphics[width=.45\textwidth]{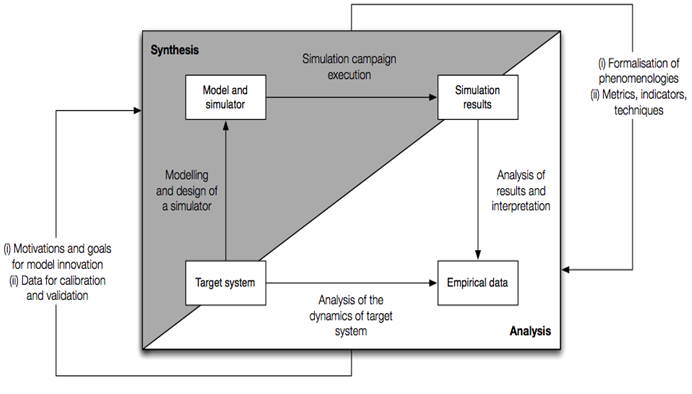}
\centering
\caption{An integrated cycle of pedestrian and crowd dynamics synthesis and analysis}
   \label{fig:integrated}
\end{figure}

Crowd studies represent successful applications of researches carried out in the context of computer simulation and computer vision. In fact, comprehensive studies require the synthesis of pedestrians and crowd behaviour but the developed models must be (i) properly calibrated and validated by means of data acquired on the field and (ii) informed by the specific contextual conditions of the simulated environment (e.g. number and positions of pedestrians in the area). Synthesis requires thus the results of analysis. In turn, the analysis of crowding phenomena can benefit from results on the side of synthesis: researches on the latter often produce formalization of phenomena, lead to the definition of metrics and indicators to evaluate the generated dynamics. These concepts and mechanisms can represent a useful contribution towards the automation of the analysis techniques that, thanks to the development of computer vision techniques, can actually produce useful information even from cluttered scenes like those taken from security cameras in public spaces. The overall resulting cycle of integrated synthesis and analysis of pedestrian and crowd dynamics is depicted in Figure~\ref{fig:integrated}.

\begin{figure}[t]
\centering
\includegraphics[width=.45\textwidth]{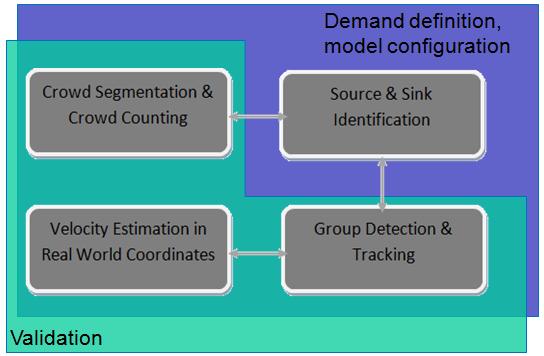}
\caption{Integration of analysis and synthesis}
\label{fig:integrate}       
\end{figure}

\subsection{Crowd Synthesis For Crowd Analysis}

The main problem that crowd synthesis and crowd anslysis share is the validation of the results. For example, in order to evaluate the accuracy of people counting algorithm, it is necessary to manually count the people in the scene, which is time consuming job. For pedestrian tracking algorithm, one must assign a label to each pedestrian and track it through thousand of frames, which is again a tedious job. Moreover, computer vision algorithms that rely on traning data require a considerable amount of training data, which is very difficult to obtain.
Next, we present some approaches where crowd simulation algorithms either provide training data or validate computer vision algorithms.

~\cite{andrade2005simulation} proposed a method which generates a video that simulates dangerous situations in crowded scenes for example, blocked exit, panic, congestion etc. We can find very rare visual evidence of these kind of scenarios and usually it is very risky to reproduce such events in a controllable way. Therefore, computer simulations of such scenarios helps to provide training data for computer vision algorithms that can applied to crowd monitoring.

~\cite{allain2009crowd} proposed an algorithm to segment and characterize crowd flows with optimal control theory. In order to validate the results, they used a crowd simulation algorithm to generate ground truth data.

Existing crowd simulation techniques, despite providing support to computer vision algorithms in terms of providing training data and validating the results, suffer from limitations. Crowd simulation algorithms are purely based on mathematical models that can simulate the known behaviour of a person but could not simulate unpredictable behaviours. Moreover, it is hard to simulate high density crowd that incorporates collision avoidance.

\subsection{Crowd Analysis For Crowd Synthesis}

As discussed above, crowd simulation can provide support to crowd analysis, in the same way, the reverse is also true. For example, it is desirable for a crowd simulatior to mimic real crowds in the virtual world. In order to achieve this, crowd simulator should reflect the physical changes in the environment,i.e, flow of people, velocity of pedestrians, density etc.~\cite{courty2007crowd} proposed a framework which captures information realted to the movement of the people using optical flow and feed a crowd simulator. They generate velocity fields over time to simulate the realistic behaviour of real crowd. In~\cite{musse2007using,paravisi2008continuum} proposed approaches for capturuing motion information from the real world by detecting and tracking individuals. The trajectories belonging to different motion patterns are clustered into different classes. Velocity fields are generated on the top of acheived clusters. This information is later on feeded to the crowd simulator.

It is important to mention that quantifying the reaslism of the crowd simulation results is still an open problem. For quantifying the realism, we must compare the parameters of the real crowd movement (such as main direction, speed and crowd density) to the simulated one.

In this paper, we propose a computer vision tool set (which is a blend of different frameworks) for a crowd that computes several measurements and provide useful information to feed a crowd simulator for initialization and validation.\ref{fig:integrated} illustrates the infrastructure of proposed toolset.

\section{Crowd Flow Segmentation and Crowd Counting}
\label{sec:2}

In this section, we discuss an important contribution that this tool set can give to the pedestrian and crowd safety is to localise large and conflicting flows in crowds. Such kind of conflicting motion patterns may lead to the congestions which ultimatley ends with crowd desasters. Therefore, early detection of such kind of motion patterns and more importantly, counting the number of people in these situations are important steps for decision makers and crowd managers. The purpose of crowd flow segmentation is to locate those groups in crowds that are distinct and spatio-temporally dominant. In order to achieve the goal of crowd flow segmentation, we generate global representation of the scene by localizing all the distinct regions/segments in the scene. we extract global representation of the scene by computing dense optical flow that computes a change at every pixel. such kind of global representation of scene makes us independent of detection and tracking of individuals in the scene. Detection and tracking pedestrians are the traditional methods for crowd analysis. But these traditional methods works well in low density situations but robustness of these methods becomes very low when applied to high density scenarios.

In high density situations, the researchers usually extract global information from the scene by using optical flow. Like \cite{ali2007lagrangian} proposed a dynamical system for crowd flow segmentation by detecting  lagrangian coherent structures in the phase space.\cite{ozturk2010detecting} detects dominant flows by detecting and tracking of SIFT features. \cite{eibl2008evaluation} employed a spectral clustering for crowd flow segmentation by computing sparse optical flow.\cite{srivastava2011crowd} extract multiple visual features for crowd flow estimation. We observe that after flow segmentation, the above methods fail to detect small flows and unclear boundaries among different flows. Moreover, the above methods are computally expensive and can not be applicable in real time. A relatively fast method is proposed in \cite{li2012crowd}, where crowd flow is segmented by using derivative curve of the histogram of angle matrix. Since this methods considers only the peaks of the histogram curve, therefore it loses a lot of meaningful information about the crowd flows. In order to caputre whole motion information in the scene, we compute dense optical flow followed by the K-means clustering. After K-means clustering small blobs appears at the boundaries of distinct flows, which is removed by our blob absorption approach. Comparing to the state-of-the-art methods, our methods can detect small as well as large flows and by employing blob absorption approach, we detect clear boundaries among distinct flows. After segmentation, we cound the number of people in each segment. The overall framework is presented in the next section.

\section{Crowd Flow Segmentation and Crowd Counting}
\label{sec:2}

In this section, we discuss an important contribution that this tool set can give to the pedestrian and crowd safety is to localise large and conflicting flows in crowds. Such kind of conflicting motion patterns may lead to the congestions which ultimatley ends with crowd desasters. Therefore, early detection of such kind of motion patterns and more importantly, counting the number of people in these situations are important steps for decision makers and crowd managers. The purpose of crowd flow segmentation is to locate those groups in crowds that are distinct and spatio-temporally dominant. In order to achieve the goal of crowd flow segmentation, we generate global representation of the scene by localizing all the distinct regions/segments in the scene. we extract global representation of the scene by computing dense optical flow that computes a change at every pixel. such kind of global representation of scene makes us independent of detection and tracking of individuals in the scene. Detection and tracking pedestrians are the traditional methods for crowd analysis. But these traditional methods works well in low density situations but robustness of these methods becomes very low when applied to high density scenarios.

In high density situations, the researchers usually extract global information from the scene by using optical flow. Like \cite{ali2007lagrangian} proposed a dynamical system for crowd flow segmentation by detecting  lagrangian coherent structures in the phase space.\cite{ozturk2010detecting} detects dominant flows by detecting and tracking of SIFT features. \cite{eibl2008evaluation} employed a spectral clustering for crowd flow segmentation by computing sparse optical flow.\cite{srivastava2011crowd} extract multiple visual features for crowd flow estimation. We observe that after flow segmentation, the above methods fail to detect small flows and unclear boundaries among different flows. Moreover, the above methods are computally expensive and can not be applicable in real time. A relatively fast method is proposed in \cite{li2012crowd}, where crowd flow is segmented by using derivative curve of the histogram of angle matrix. Since this methods considers only the peaks of the histogram curve, therefore it loses a lot of meaningful information about the crowd flows. In order to caputre whole motion information in the scene, we compute dense optical flow followed by the K-means clustering. After K-means clustering small blobs appears at the boundaries of distinct flows, which is removed by our blob absorption approach. Comparing to the state-of-the-art methods, our methods can detect small as well as large flows and by employing blob absorption approach, we detect clear boundaries among distinct flows. After segmentation, we cound the number of people in each segment. The overall framework is presented in the next section.

\subsubsection{Foreground Extraction}

Extracting foreground objects is the most important pre-processing step and therefore forms the basis of our framework. Foreground extraction is useful for detection, tracking and understanding the behavior of the object. A survey on motion detection techniques can be found in \cite{moeslund2001survey}, In traditional visual surveillance, usually with a fixed camera, research use background subtraction methods, where the moving objects are detected, if the intensity values of pixles in the current frame deviate significantly from the background. These kind of methods are prone to noise, a small change in illumination can be detected as foreground. We use adaptive Gaussian mixture model (GMM) to generate a foreground mask, $f_{g(x,y,t)}$,  which is more robust to these kind of noises. GMM is very good in separating the foreground objects from the background but it can not compute the change in every pixel of the image. Usually crowded objects move in wide areas, and for flow segmentation problem, we need to detect change in every pixel. Therefore, instead of using the foreground mask generate by GMM, we generate another foreground mask $ f_{hs(x,y,t)}$ by computing the dense optical flow, smoothed by gaussian and median filters. We use Horn and Schunk (HS), but any method for dense optical flow computation can be used. Since we compute flow vector at each pixel, so each pixel has the magnitude and direction values. we use magnitude information of the flow vector to generate teh foreground mask, all the pixels which have higher magnitude than a predefined threshold will be classified as foreground. Direction information of flow vectors can be used in crowd flow segmentation, since we are segmenting the flows on the basis of orientations. Optimal foreground mask $f_{out(x,y,t}$ is obtained by logical product of $ f_{hs(x,y,t)}$ and $f_{g(x,y,t)}$. Later on, we apply morphological processes like morphological opening and closing on the $f_{out(x,y,t}$. Morphological process smooths the section of contours, eliminates small holes and fills gaps in contours. Segmentation block  segments the crowd flows into different clusters, $C'_{j(x,y,t)}$, by employing $K$-means clustering followed by blob absorption method. To estimate the number of people in each flow segment, we take logical product of each cluster $C'_{j(x,y,t)}$ and foreground mask $f_{out(x,y,t)}$ and count the number of people by blob analysis and blob size optimization methods.

\subsubsection{Crowd Flow Segmentation}

After foreground extraction, the next step is to compute motion flow field. Motion flow field is a set of independent flow vectors and each flow vector is associated with its spatial location and its orientation. Since we compute motion flow field at every frame, therefore we termed it \textit{Instantaneous flow field}, which captures temporal information of the motion patterns and can be used to learn motion patterns in the video. Consider a feature point $i$, its flow vector $F_i$ is represented by its location $X_i$ and its velocity $V_i$, i.e , $F_i$ = $\lbrace X_i, V_i \rbrace$. Since we are computing the flow vector of each feature point that belongs to the foreground objects, therefore, instantaneous motion field is given by $ \lbrace F_1,F_2,...F_n \rbrace $. This motion flow field is a n x 4 matrix, where each row of matrix represents the feature point $i$ and column represents its corresponding spatial location and velocity. Each flow vector represents a motion in a specific direction, therefore, we can not infer any meaningful information about the dominant flows from motion flow field alone. For detecting dominant motion patterns, we need to compute similarity among flow vectors and cluster them into multiple groups. In order to cluster similar flow vectors, we employ $K-means$ clustering algorithm. This process of grouping flow vectors into distinct groups is called flow segmentation.

\begin{figure}[t]
\centering
\includegraphics[width=.45\textwidth]{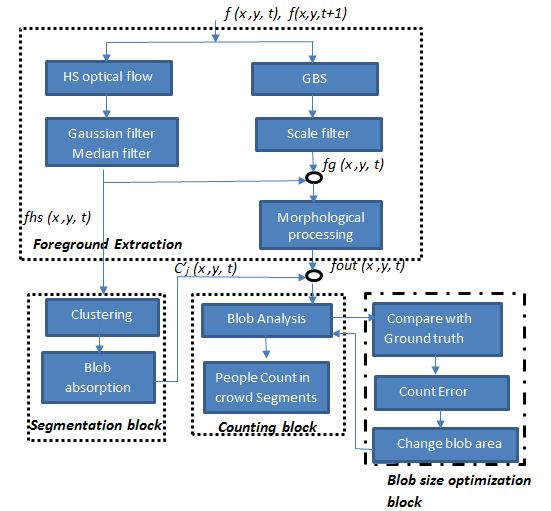}
\caption{Crowd flow segmentation and counting}
\label{fig:framework}       
\end{figure}

\begin{figure}[t]
\raggedright
\includegraphics[width=.97\columnwidth]{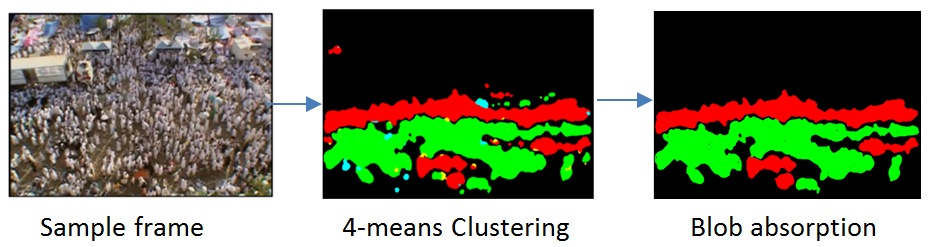}
 \caption{Results of 4-means clustering and blob absorption in a Hajj video frame}
   \label{fig:kmeans}
\end{figure}

From our experiments, we observe that after $K-means$ clustering, small blobs appear, these small blobs represents small clusters and detected due to (1), if the object move slowly, the central vectors of pixels are not same as the vectors at the boundaries, and hence clustered into two different groups. (2) the optial flow vectors at the boundaries of two opposite is always ambigious, and as a result small clusters appear at the boundaries of two opposite flows. In order to get rid off such kind of noisy clusters, we adopt a blob absorption approach which works on the principle of \textit{Big fish eats small fishes}, where small blobs are absorbed either by a big cluster or by background. In our proposed blob absorption approach, background is also assumed as a big cluster, since in natural images, large amount of pixels correspond to background. A detailed description of blob absorption approach is given in \cite{khan2014detecting}. Figure~\ref{fig:kmeans} shows a sample frame from a hajj video sequence, where the people are moving in two dominant directions. After employing K-means clustering (K=4,in this case), crowd is segmented into two main dominant flows with small clusters at the boundaries which are removed after employing blob absorption. After blob absorption, the crowd is segmented into dominant flows with the obvious boundaries. Now, the crowd is segmented into different segments. The next step is to count the number of people in each segment.

\subsubsection{Crowd Counting}

In this section, we describe the methodology for counting
the people in each segment. In low density crowds, where people are spread sparsely in the environment and each individual is clearly visible, we can use traditional methods of human detection and tracking to count the number of people. Therefore, it implies that in low density situations, counting people is a trivial job. whereas in high density situations, where the people in the environment are tight packed, highly occluded and due to less number of pixels per person, it is extremely challenging to detect the people using a human descriptor and hence it makes the counting problem even more difficult. Therefore, as a solution, we perform global analysis by employing blob analysis and blob size optimization techniques on foreground image and estimate the number of people in high density crowds.

Blob analysis is a technique that computes statistics for blobs in an image. Blobs are the connected regions in the binary image and usually represent the moving objects in the scene. Since there are many blobs of different sizes representing different moving objects in an image, we need to find out optimum size of the blob that can serve as a threshold. The blobs with size/area above the specified threshold will not be considered (for instance, when counting pedestrians in road videos, these large blobs might be related to cars). For computing optimum size of the blob, we employ blob size optimization algorithm in~\cite{khan2014detecting} and~\cite{arif2013counting}. In our experimental set up, in order to find the optimum blob size, we use four or five frames of video selected randomly. For each of the selected frame, we compute optimum size and final optimum blob size $A'$ is the mean of all four or five blob sizes. We use $A'$ for counting people in rest of video frames.

\section{Crowd Behavior: Identifying Sources and Sinks}

Crowded scenes are composed of large number of people. The people in the crowd exhibits different behaviours and understanding crowd behaviours without analysing the actions of individuals (in crowds) are always advantangeous to  designers, planners and decision makers. Automatic detection of crowd behaviours have many applications, such as prediction of congestion which lead to unnecessary delays, detection of abnormal events which lead to crowd disasters. Crowd behaviour modeling and understanding has important pre-processing task (i) extracting spatio-temporal motion information (e.g, trajectories), (ii) identification of source(entry) and sink(exit) points of trajectories, (iii) interaction of trajectories. We can not extract sptio-temporal motion information(trajectories) by empolying crowd flow segmentation framework discussed in section \ref{sec:2}, therefore in order to automate the process of crowd behaviour understanding, we devise a new framework by adopting two novel algorithms, the first able to generate long, dense, reliable and accurate pedestrian trajectories and the second clustering them into dominant flows. The final global flows not only provide direct information about the characterization of flows but also provide a starting point for the further high level analysis of crowd behaviour.

\begin{figure}[t]
\center
\includegraphics[width=.97\columnwidth]{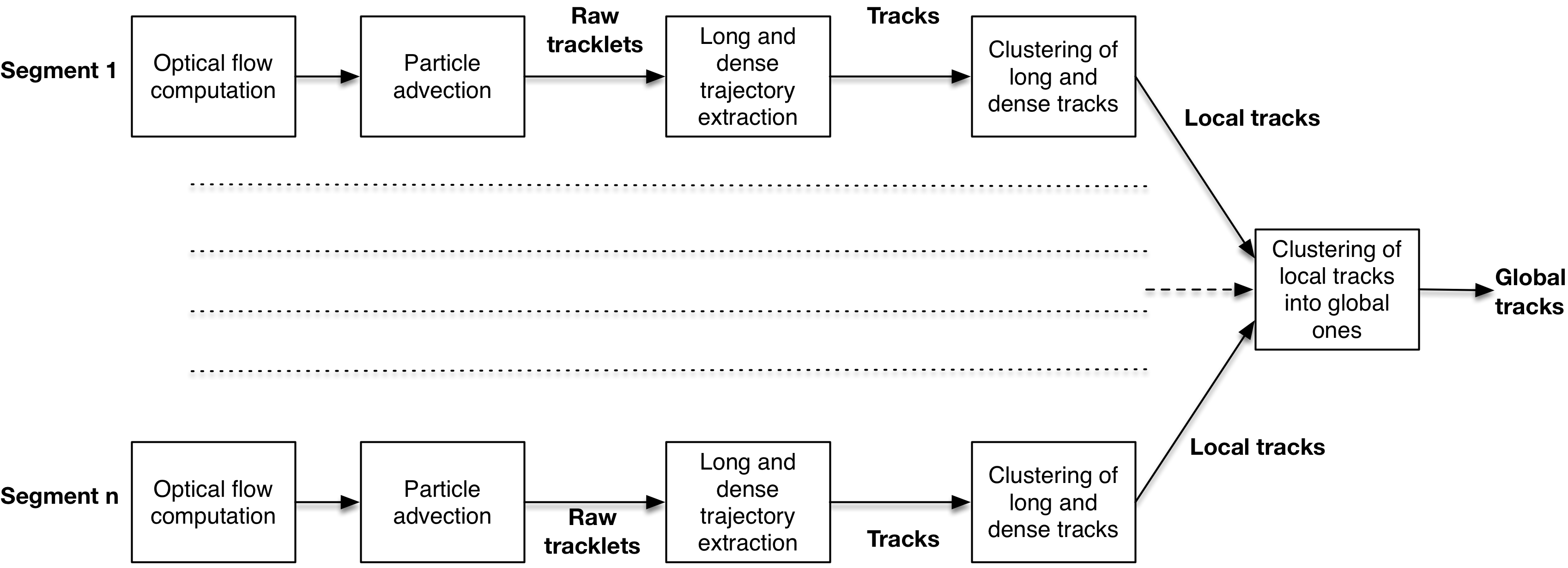}
\caption{Source and sink identification framework}
\label{fig:source-sink}       
\end{figure}

The approach starts by dividing the input video into multiple segments of equal length. The first frame of each video segment is overlaid by grid of particles initializing a dynamical system defined by optical flow as in~\cite{solmaz2012identifying}. Particle trajectories are extracted by integrating the dynamical system over time. We identify sources, sinks and characterize main flows by analysing particle trajectories using unsupervised hierarchical clustering algorithm, where similarity among the trajectories is measured by Longest Common Sub-Sequence (LCSS) metric. The final global tracks are achieved by clustering local tracks through the same clustering algorithm. The above steps of the framework is illustrated in Figure~\ref{fig:source-sink} and described in the following section.

\subsection{Achieving Reliable Trajectories}

As mentioned above, the input to our framework is a sequence of video frames which is automatically divided into $n$ of segments, each of size $k$ frames. Since it is extremely hard to detect and track pedestrians in high density situations, therefore, we rely global analysis by employing optical flow.

\subsubsection{Particle Advection}

\begin{figure}[t]
\includegraphics[width=.97\columnwidth]{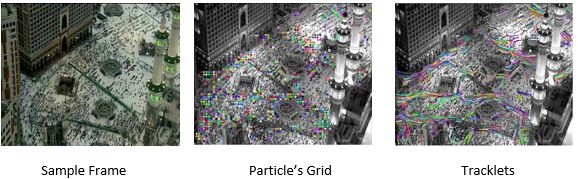}
\caption{First image is the sample frame. Second image are the source points of particles while the third image are the particle trajectories obtained after the advection process}
\label{fig:advec}       
\end{figure}

In order to extract trajectories, we compute dense optical flow between two consecutive frames of every segment. We then initialize a continuous dynamical system by overlaying grid of particles on the initial optical flow field of the video segment and each position of the particle represent the source point.
After initiliazing the grid of paritcles, the next step is to advect the particles in forward time over the optical flow field. As a result of particle advection, small duration trajectories called $tracklets$ are obtained as shown in Figure~\ref{fig:advec}.

\subsubsection{Achieving Final Flows Through Clustering}

Tracklets achieved through the particle advection are short duration and therefore have a limited spatial extent. These tracklets fail to represent important characteristics of the overall motion. Moreover, they proivide inadequate information that could help in identitying the source and sink points of the dominant flows and hence inappropriate to directly applied for behaviour understanding. In order to achieve better representation of motion information, we need to cluster these tracklets into longer trajectories. This becomes a combinatorial matching problem that we define and solve recursively.

Our proposed clustering algorithm is based on the assumption that the tracklets corresponding to single motion pattern are similar in orientation but the sources and sinks are spatially different. These tracklets start and end at different locations but the sink of one tracklet lie spatially close to the source of other tracklet. The algorithm in~\cite{khan2015analyzing} exploits spatial closeness of source and sink locations and  similarity among the tracklets by combining them into longer tracks.

After achieving long tracks, the next step is to cluster similar tracks into local tracks by employing hierarchical clustering algorithm using the following procedure.

\begin{enumerate}

\item We sort the tracks in descending order on the basis of their length. We compute the length of track as euclidean distance between its start and end point. Let $L_T$ is the sorted list of tracks.

\item We also set up a list of clusters $L_C$, initially containing  one cluster associated to the first track $T_1$ (the longest one) in the list  and it is considered as an initial cluster center.

\item We select bottom most track from the list, $T_s$, and compare it with the centers of all clusters present in $L_C$ using longest common sub-sequence metric. If similarity value is greater than a threshold, then tracklet $T_s$ is assigned to the current cluster, otherwise, we initialize a new cluster with the center $T_s$. We delete the tracklet $T_s$ from the list $L_T$ after assignment to a cluster.

\item  If the cluster's size exceeds a positive value of $S$, then we update the cluster center by using $K^{th}$ order least square polynomial regression.  We use $S = 30$ in our experiments.

\item We repeat the previous step until $L_T$ is not empty.

\end{enumerate}

\section{Group Detection in Crowds}

In the previous frameworks, we tried to tackle crowd as whole entity where individuality does not exist. We replicate this assumption from a $\it{Popular}\it{ Mind}\it{ Theory}$ in~\cite{le1897crowd}, where the crowd is defined as "pathological monster with no individual consciousness". However, this theory are applicable to extremely high density crowds. But in relatively low density situations, for example, airport, stations, shopping mall and concerts, where most of the people tend to move in groups. The members of the groups tend to maintain spatial and temporal correlations and therefore behavior of the crowd is usually influenced by these social relationships. Realizing the importance of group behavior and its influence on crowd, we propose an approach for automatic detection and tracking of groups in crowds.

\begin{figure}[t]
\includegraphics[width=0.50\textwidth]{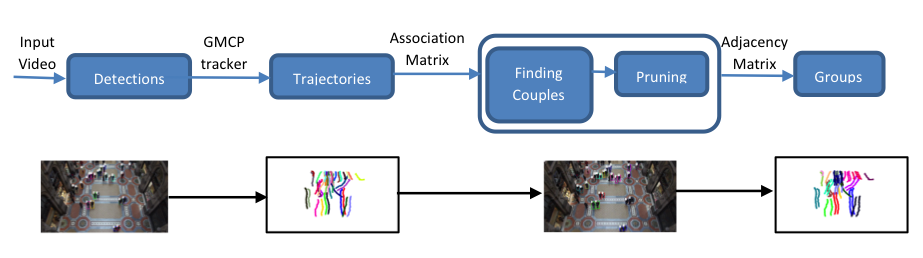}
\centering
\caption{Proposed Methodology for Group Detection.}
   \label{fig:propose-method}
\end{figure}

The proposed approach starts by detecting individual pedestrians in video frame and then track the detected pedestrian through multiple frames using GMCP tracker by~\cite{zamir2012gmcp}. The trajectory of pedestrian is a set of tuples $(x,y,t)$, where $x$ and $y$ are the horizontal and vertical coordinates of the location at time $t$. We then define an \emph{Association Matrix}, which captures the joint probability distribution of source and sink locations of all pedestrian trajectories. In order to capture the probability distributions of source and sink locations of trajectories, we assume two discrete random variables $\bf{X}$, representing ``source'' locations of the trajectories and $\bf{Y}$ representing ``sink'' locations.

An \emph{Association Matrix}
 for $n$ trajectories is shown below. 
\[
P(X,Y) = \begin{Bmatrix}
    p_{11} & p_{12} & p_{13} & \dots  & p_{1n} \\
    p_{21} & p_{22} & p_{23} & \dots  & p_{2n} \\
    \vdots & \vdots & \vdots & \ddots & \vdots \\
    p_{n1} & p_{n2} & p_{n3} & \dots  & p_{nn}
 \end{Bmatrix}
\]

Each element of \emph{Association Matrix} shows the probability distribution of source and sink location of a single pedestrian $k$ over all other $n$ pedestrians. Association matrix captures the walking behavior of a pedestrian relative to other pedestrians in the scene. A single pedestrian who is not a member of any group tends to stop or move freely in the environment. Moreover, he tends to keep his distance from other pedestrians. In the same way, members of the group tend to maintain small proximity within members and large with other individuals. This type of behavior uniquely identify the group which can be captured by the association matrix.

A three step bottom-up hierarchical clustering approach is employed to discover couples. In the first step, we assign distinct cluster identifiers by treating each pedestrian as a separate cluster. In the second step, the algorithm adopts a greedy approach to find a best possible member for each pedestrian to form a couple. The algorithm groups two pedestrians in a group by measuring the difference between their probability distribution by using \emph{Kullback-Leibler (KL) divergence}, also known as relative entropy, denoted by $D_{KL}(P_r||P_k)$. If $KL$ value is more than a predefined threshold, then pedestrians are termed as bad couples. After pruning of bad couples, groups are discovered using \emph{Adjacency Matrix}, which captures the connectivity information among all pedestrians. Figure~\ref{fig:group_examples} shows qualitative results of the proposed framework.

\begin{figure}[t]
\centering
\subfigure[Hotel Sequence]{\includegraphics[width=0.41\textwidth]{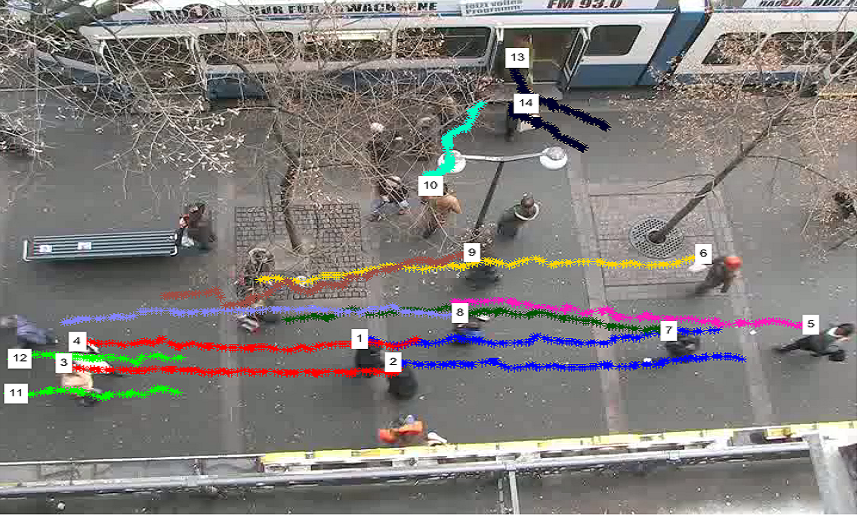}}
\subfigure[Gallery Sequence]{\includegraphics[width=0.46\textwidth]{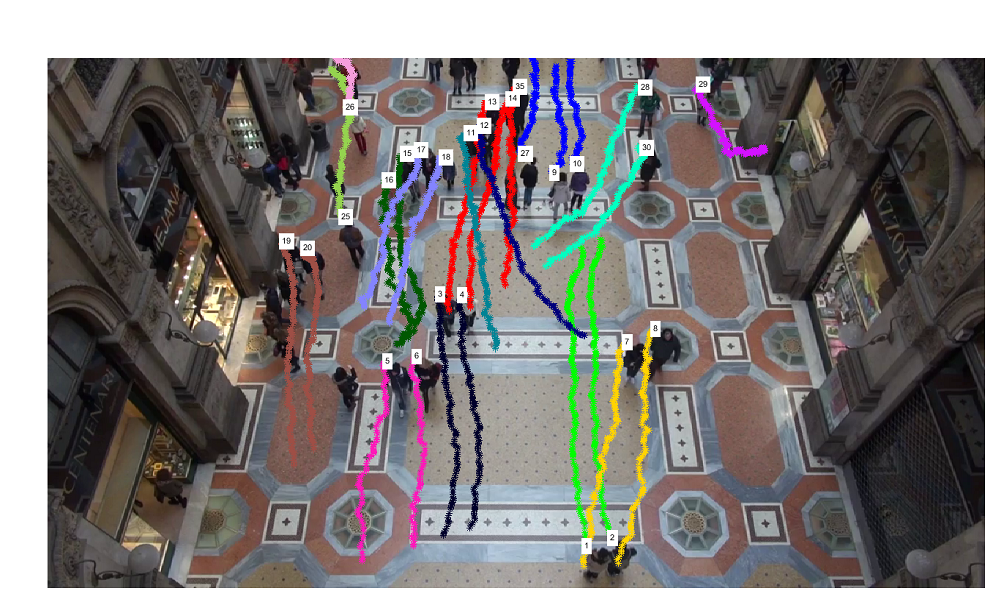}}
\caption{Qualitative results of different video sequences}
\end{figure} \label{fig:group_examples}

\section{Application of Computer Vision Toolset}

\begin{figure}[t]
\includegraphics[width=0.41\textwidth]{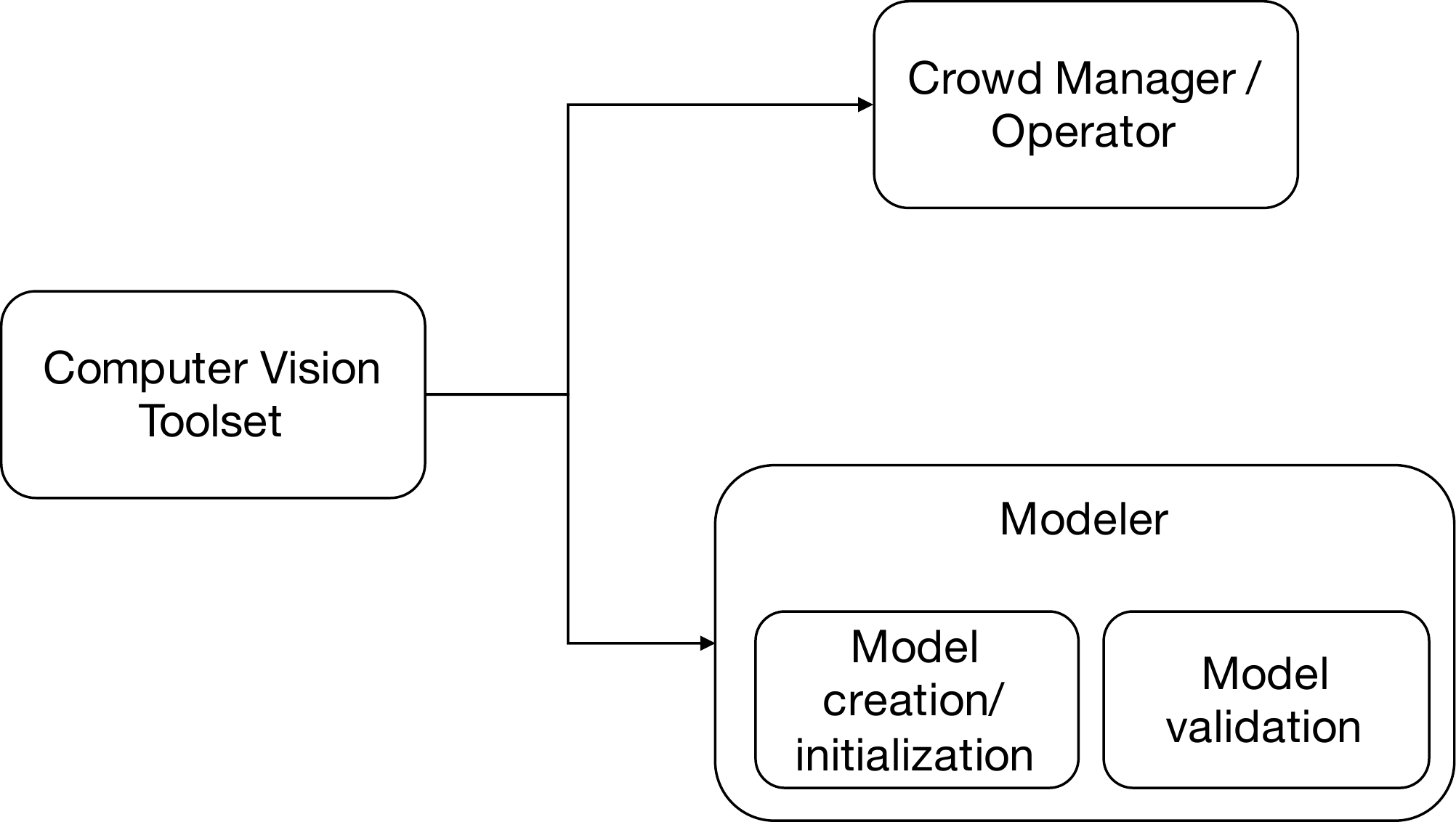}
\caption{Application of the proposed tool set}
\label{fig:toolset}       
\end{figure}

The proposed computer vision toolset has two fold applications, one to crowd manager and other to the modeler as shown in Figure~\ref{fig:toolset}. Due to to the complex dynamics of the crowd, crowd management is becoming a daunting job for the crowd managers and security staff. In such high density crowded situations, to ensure the safety of people, low cost surveillance cameras are  installed at different locations that can cover the whole crowd. The analysts sitting in the surveillance room watching over multiple Tv screens in order to detect some abnormal events. Such manual analysis of crowd is a tedious job and usually prone errors. These problems necessitate the development of methods and tools that can automatically analyze the crowd and can give reliable estimate about the density and detect specific activities. The proposed toolset can provide information about the crowd size, distinct motion patterns, crowd behaviors and pedestrian groups detection.

During the last 15 years, many crowd and pedestrian simulation models have been proposed in literature~\cite{thalmann2007crowd}. Simulation models has been providing a support in decision-making since last many years. These simulation models mimic the real-world crowds but they never exactly imitate the real world system. Therefore every simulation model requires empirical evidences for the validation. For example, in several crowd simulation models, it is desirable to model a real scenario, and in order to study how the physical environment would affect the flow of the people. In this case, it is important to achieve information about the motion of people, density, dominant directions, velocity. The manual extraction of this kind of information is very tedious, time consuming and usually prone to errors. This motivates the use of proposed computer vision tool set that can provide useful information that could be helpful in initial configuration of simulation models and also provide support in the validation phase.

\begin{figure}[t]
\includegraphics[width=0.41\textwidth]{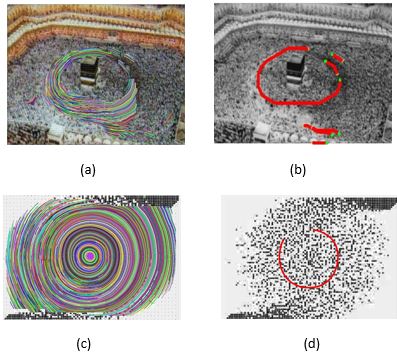}
\caption{(a) trajectories extraced during the particle advection process on real time video, (b) final circular flow in real time video, (c) trajectories extracted through particle advection process in simulated video, (d) final circular flow in simulated video}
\label{fig:example}       
\end{figure}

In Figure~\ref{fig:example}, we presented an example where we capture information from the real time video, where the people are circulating around the Kaaba performing a religious ritual. In this example, we capture movements of the people by employing our method discussed in section~\ref{sec:2}, where a optical flow based dynamical system is initiated followed by the particle advection. As result of advection process, trajectories are obtained as shown in Figure~\ref{fig:example}(a). The acheieved trajectories are clustered into a single dominant and coherent flow as shown in Figure~\ref{fig:example} (b). This information is feeded to the simulator in order to reproduce spiral movements of the people around a central object. The pedestrian motion is modelled by using cellular automata (CA). A more realistic behaviour is obtained by incorporating floor field to the wall avoidance and lane formation as in~\cite{shimura2016simulation}. In order to validate the circular movement of the people, we also initialize an optical flow base dynamical system followed by particle advection for the simulated video as shown in~Figure~\ref{fig:example}(c). Trajectories achieved after applying particle advection process on a simulated video, are clustered into a single flow as shwon in Figure~\ref{fig:example}(d). The circular flow detected in simulated video highlights the fact that the spiral movements of pedestrians are accurately modelled.

\bibliographystyle{IEEEtran}
\bibliography{mybibfile}

\begin{thebibliography}{10}
\providecommand{\url}[1]{#1}
\csname url@samestyle\endcsname
\providecommand{\newblock}{\relax}
\providecommand{\bibinfo}[2]{#2}
\providecommand{\BIBentrySTDinterwordspacing}{\spaceskip=0pt\relax}
\providecommand{\BIBentryALTinterwordstretchfactor}{4}
\providecommand{\BIBentryALTinterwordspacing}{\spaceskip=\fontdimen2\font plus
\BIBentryALTinterwordstretchfactor\fontdimen3\font minus
  \fontdimen4\font\relax}
\providecommand{\BIBforeignlanguage}[2]{{%
\expandafter\ifx\csname l@#1\endcsname\relax
\typeout{** WARNING: IEEEtran.bst: No hyphenation pattern has been}%
\typeout{** loaded for the language `#1'. Using the pattern for}%
\typeout{** the default language instead.}%
\else
\language=\csname l@#1\endcsname
\fi
#2}}
\providecommand{\BIBdecl}{\relax}
\BIBdecl

\bibitem{andrade2005simulation}
E.~L. Andrade and R.~B. Fisher, ``Simulation of crowd problems for computer
  vision,'' in \emph{First International Workshop on Crowd Simulation}, vol.~3,
  2005, pp. 71--80.

\bibitem{allain2009crowd}
P.~Allain, N.~Courty, and T.~Corpetti, ``Crowd flow characterization with
  optimal control theory,'' in \emph{Computer Vision--ACCV 2009}.\hskip 1em
  plus 0.5em minus 0.4em\relax Springer, 2009, pp. 279--290.

\bibitem{courty2007crowd}
N.~Courty and T.~Corpetti, ``Crowd motion capture,'' \emph{Computer Animation
  and Virtual Worlds}, vol.~18, no. 4-5, pp. 361--370, 2007.

\bibitem{musse2007using}
S.~R. Musse, C.~R. Jung, J.~Jacques, and A.~Braun, ``Using computer vision to
  simulate the motion of virtual agents,'' \emph{Computer Animation and Virtual
  Worlds}, vol.~18, no.~2, pp. 83--93, 2007.

\bibitem{paravisi2008continuum}
M.~Paravisi, A.~Werhli, J.~Junior, R.~Rodrigues, C.~Jung, and S.~Musse,
  ``Continuum crowds with local control,'' in \emph{Computer Graphics
  International}, 2008, pp. 108--115.

\bibitem{ali2007lagrangian}
S.~Ali and M.~Shah, ``A lagrangian particle dynamics approach for crowd flow
  segmentation and stability analysis,'' in \emph{Computer Vision and Pattern
  Recognition, 2007. CVPR'07. IEEE Conference on}.\hskip 1em plus 0.5em minus
  0.4em\relax IEEE, 2007, pp. 1--6.

\bibitem{ozturk2010detecting}
O.~Ozturk, T.~Yamasaki, and K.~Aizawa, ``Detecting dominant motion flows in
  unstructured/structured crowd scenes,'' in \emph{Pattern Recognition (ICPR),
  2010 20th International Conference on}.\hskip 1em plus 0.5em minus
  0.4em\relax IEEE, 2010, pp. 3533--3536.

\bibitem{eibl2008evaluation}
G.~Eibl and N.~Br{\"a}ndle, ``Evaluation of clustering methods for finding
  dominant optical flow fields in crowded scenes,'' in \emph{Pattern
  Recognition, 2008. ICPR 2008. 19th International Conference on}.\hskip 1em
  plus 0.5em minus 0.4em\relax IEEE, 2008, pp. 1--4.

\bibitem{srivastava2011crowd}
S.~Srivastava, K.~K. Ng, and E.~J. Delp, ``Crowd flow estimation using multiple
  visual features for scenes with changing crowd densities,'' in \emph{Advanced
  Video and Signal-Based Surveillance (AVSS), 2011 8th IEEE International
  Conference on}.\hskip 1em plus 0.5em minus 0.4em\relax IEEE, 2011, pp.
  60--65.

\bibitem{li2012crowd}
W.~Li, J.-H. Ruan, and H.-A. Zha, ``Crowd movement segmentation using velocity
  field histogram curve,'' in \emph{Wavelet Analysis and Pattern Recognition
  (ICWAPR), 2012 International Conference on}.\hskip 1em plus 0.5em minus
  0.4em\relax IEEE, 2012, pp. 191--195.

\bibitem{moeslund2001survey}
T.~B. Moeslund and E.~Granum, ``A survey of computer vision-based human motion
  capture,'' \emph{Computer vision and image understanding}, vol.~81, no.~3,
  pp. 231--268, 2001.

\bibitem{khan2014detecting}
S.~D. Khan, G.~Vizzari, S.~Bandini, and S.~Basalamah, ``Detecting dominant
  motion flows and people counting in high density crowds,'' 2014.

\bibitem{arif2013counting}
M.~Arif, S.~Daud, and S.~Basalamah, ``Counting of people in the extremely dense
  crowd using genetic algorithm and blobs counting,'' \emph{IAES International
  Journal of Artificial Intelligence}, vol.~2, no.~2, p.~51, 2013.

\bibitem{solmaz2012identifying}
B.~Solmaz, B.~E. Moore, and M.~Shah, ``Identifying behaviors in crowd scenes
  using stability analysis for dynamical systems,'' \emph{Pattern Analysis and
  Machine Intelligence, IEEE Transactions on}, vol.~34, no.~10, pp. 2064--2070,
  2012.

\bibitem{khan2015analyzing}
S.~D. Khan, S.~Bandini, S.~Basalamah, and G.~Vizzari, ``Analyzing crowd
  behavior in naturalistic conditions: Identifying sources and sinks and
  characterizing main flows,'' \emph{Neurocomputing}, 2015.

\bibitem{le1897crowd}
G.~Le~Bon, \emph{The crowd: A study of the popular mind}.\hskip 1em plus 0.5em
  minus 0.4em\relax Fischer, 1897.

\bibitem{zamir2012gmcp}
A.~R. Zamir, A.~Dehghan, and M.~Shah, ``Gmcp-tracker: Global multi-object
  tracking using generalized minimum clique graphs,'' in \emph{Computer
  Vision--ECCV 2012}.\hskip 1em plus 0.5em minus 0.4em\relax Springer, 2012,
  pp. 343--356.

\bibitem{thalmann2007crowd}
D.~Thalmann, \emph{Crowd simulation}.\hskip 1em plus 0.5em minus 0.4em\relax
  Wiley Online Library, 2007.

\bibitem{shimura2016simulation}
K.~SHIMURA, S.~D. KHAN, S.~BANDINI, and K.~NISHINARI, ``Simulation and
  evaluation of spiral movement of pedestrians: Towards the tawaf simulator.''
  \emph{Journal of Cellular Automata}, vol.~11, no.~4, 2016.

\end{thebibliography}

\end{document}